\documentclass[runningheads]{llncs}
\usepackage{graphicx}
\usepackage{times}
\usepackage{latexsym}
\usepackage{mathtools}
\usepackage{multirow}
\usepackage{makecell}
\usepackage{float}
\usepackage{lipsum}
\usepackage{amsmath}
\usepackage{cleveref}
\restylefloat{table}    
\usepackage{arydshln}
\usepackage{adjustbox}
\usepackage{subfig}
\usepackage{cite}
\begin{document}

\title{Concept Identification of Directly and Indirectly Related Mentions Referring to Groups of Persons}
\titlerunning{Concept Identification of Mentions Referring to Groups of Persons}
%
\author{Anastasia Zhukova\inst{1} \and 
Felix Hamborg\inst{2,4} \and
Karsten Donnay\inst{3,4} \and
Bela Gipp\inst{1,4}}
\authorrunning{A. Zhukova et al.}
\institute{University of Wuppertal, Germany \\
\email{\{last\}@uni-wuppertal.de}\\
\url{https://dke.uni-wuppertal.de/en/} \and
University of Konstanz, Germany\\
\email{felix.hamborg@uni-konstanz.de} \and
University of Zurich, Switzerland\\
\email{donnay@ipz.uzh.ch} \and
Heidelberg Academy of Sciences and Humanities, Germany\\}

\sloppy
\maketitle
\fussy

\begin{abstract}
Unsupervised concept identification through clustering, i.e., identification of semantically related words and phrases, is a common approach to identify contextual primitives employed in various use cases, e.g., text dimension reduction, i.e., replace words with the concepts to reduce the vocabulary size, summarization, and named entity resolution. We demonstrate the first results of an unsupervised approach for the identification of groups of persons as actors extracted from a set of related articles. Specifically, the approach clusters mentions of groups of persons that act as non-named entity actors in the texts, e.g., ``migrant families'' $=$ ``asylum-seekers.'' Compared to our baseline, the approach keeps the mentions of the geopolitical entities separated, e.g., ``Iran leaders'' $\ne$ ``European leaders,'' and clusters (in)directly related mentions with diverse wording, e.g., ``American officials'' $=$ ``Trump Administration.''\footnote{\textbf{The final authenticated version is available online at \url{https://doi.org/10.1007/978-3-030-71292-1\_40}}}

\end{abstract}

\section{Introduction}
Methods for \textit{concept identification} seek to identify words and phrases that refer to the same semantic concept. As such, concept identification is a crucial task employed in various use cases, such as information summarization, information extraction, named entity resolution, and coreference resolution. While in some domains, e.g., medicine, semantic (dis)similarities are clearly distinct, in others, e.g., the news domain, phrases referring to groups of persons are often \textit{semantically highly related yet conceptually different}, e.g., ``American officials'' and ``Israeli officials'' have similar roles but act as different actors.  Identification of conceptually fine-grained groups of persons is a challenging task due to two key issues: first, high semantic relatedness of mentions that yet perform conceptually different roles, e.g., ``immigration lawyers'' and ''undocumented immigrants.'' Second, event-specific coreferential relations are often prone to high lexical diversity due to the word choice and labeling \cite{Hamborg2019}, e.g., ``Dreamers'' and ``DACA recipients.'' 

In this work, we propose an unsupervised concept identification approach that automatically extracts conceptually fine-grained clusters of related mentions referring to groups of people from a set of text documents. We narrow down our problem statement to news articles since word choice is especially subtle and rich in the news domain. The goal of our approach is to extract from news stories those group-actors that are the main content elements and yet missed by current coreference resolution and named entity recognition.  

\section{Related Work}
Concept identification is a technique important across various use cases, e.g., for dimension reduction (cf.\cite{kim2017bag, JIA2018691, chen2018anchorviz}), information extraction (cf.\cite{han2017automatic}), information summarization (cf.\cite{cambria2018senticnet}), coreference resolution of the mentions referring to the same entities (cf.\cite{subramanian-roth-2019-improving}), taxonomy construction (cf.\cite{cha2017language}), and named entity or domain concept recognition (cf. \cite{nikfarjam2015pharmacovigilance, si2019enhancing}). 

Scholars have proposed supervised tasks where a model is trained to identify domain-specific concepts, e.g., reactions to drugs \cite{si2019enhancing, nikfarjam2015pharmacovigilance}, by automatically labeling phrases with their respective concepts, e.g., persons or other named entities. Most frequently, concept identification is an unsupervised task to explore the relations between the words or phrases contained in a text \cite{JIA2018691, han2017automatic, nikfarjam2015pharmacovigilance, kim2017bag}.  Unsupervised methods use clustering, e.g., K-means \cite{JIA2018691}, which find patterns between the elements without prior knowledge. Such methods are typically integrated as preprocessing or intermediate steps so that their results can be used in downstream analysis steps. While less bound to the content of text datasets, clustering-based methods are more difficult to use because one has to find a clustering parameter configuration to yield suitable results for the dataset at hand.

\section{Methodology}

We propose an unsupervised clustering approach that identifies mentions \textit{directly referring} to the same group of individuals in a given context, e.g., ``asylum-seekers'' and ``Central American immigrants,'' and groups of individuals semantically related to countries or organizations as the representatives of both, i.e., \textit{indirectly coreferential}, e.g., ``White House officials'' -- ``Trump administration.'' For the clustering itself, we employ the core principle of two clustering algorithms: 1) OPTICS clustering algorithm \cite{optics},  i.e., we form clusters by decreasing cluster density; 2) hierarchical clustering (HC) \cite{murtagh2012algorithms}, i.e., we use the weighted average linkage criterion to merge clusters.

\subsection{Mention extraction}
\label{sec:extraction}
A \textit{mention} is a noun phrase (NP) automatically extracted from a parsed text, e.g., by CoreNLP  \cite{manning-EtAl:2014:P14-5}. We extract NPs not larger than 20 words. For each mention we assign a \textit{representative phrase} (RP), i.e., a shortened version of the phrase that includes only the most frequent dependency parsing components of a NP: heads of NPs, compounds, and adjectival and noun modifiers. We use unique RPs as clustering units, i.e., we assume that within a narrow article-based context identical RPs of different mentions $m_i$ share same meaning $rp_l = rp(m_i)$.

\begin{figure}[h]
\centering
\includegraphics[width=0.55\textwidth]{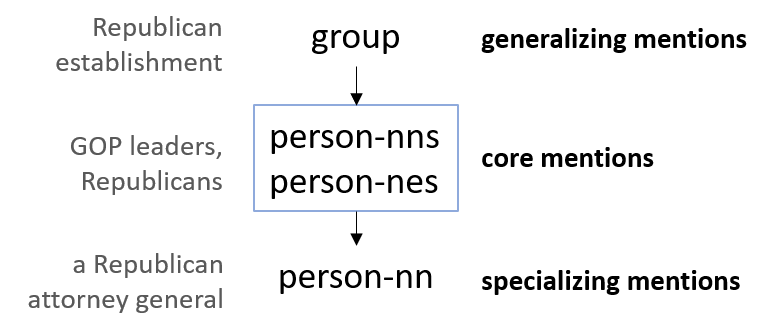}
\caption{ Level of details among the mention types. } 
\label{img:mentions}
\end{figure}

To select mentions referring to groups of persons, we apply the entity type identification methodology proposed by Hamborg et al. \cite{Hamborg2019a} and keep all mentions of four entity types: (1) multiple persons NE (``person-nes''), e.g., ``Republicans,'' (2) multiple persons non-NE (``person-nns''), e.g., ``GOP leaders,'' (3) single person non-NE (``person-nn''), e.g., ``a Republican attorney,'' and (4) group of people (``group''), e.g., ``Republican establishment.'' Fig.~\ref{img:mentions} depicts how these types form hypernym-hyponym relations. While ``group'' is the most general and aggregated type, ``person-nn'' is the type that has the largest level of details, i.e., the single instances of the groups.  
Due to the comparably balanced level of detail inherent to concepts of the types ``person-nes'' and ``person-nns,'' we coin their mentions \textit{core mentions}.

\subsection{Pipeline}
\begin{figure*}
\centering
\includegraphics[width=1.0\textwidth]{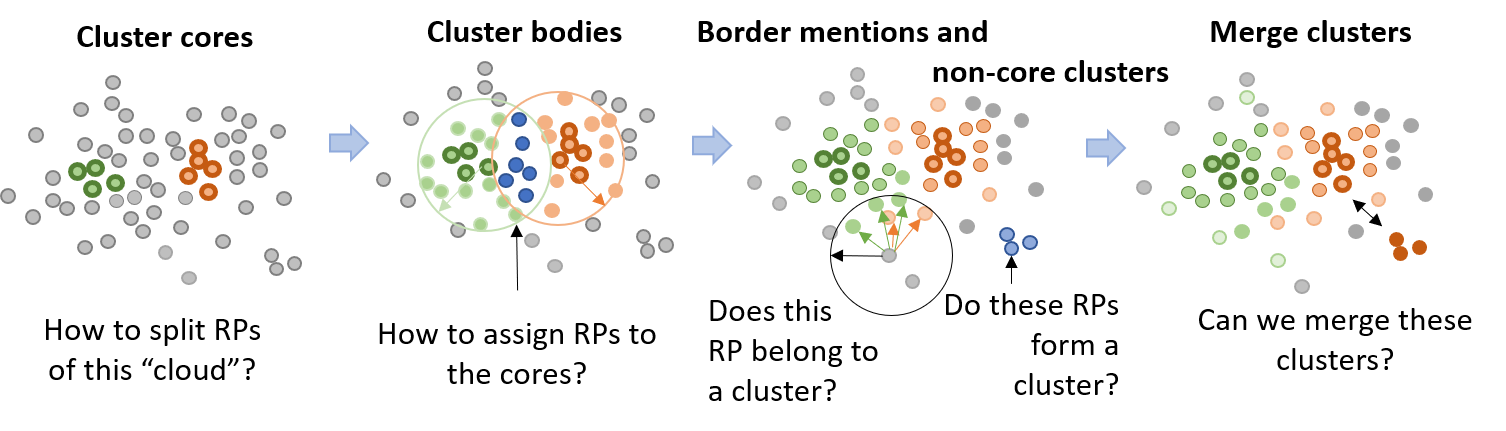}
\caption{\label{img:pipeline_graph} Identification of mention clusters.}
\end{figure*}

Our approach consists of six stages where the first identifies cluster cores and subsequent stages expand the clusters: (1) preprocessing, (2) identify cluster cores, (3) form cluster bodies, (4) add border mentions, (5) form non-core clusters, and (6) merge final clusters. Fig.~\ref{img:pipeline_graph} depicts the principle of the approach.  

\subsection{Preprocessing}
\label{sec:prep}
In early experiments, we observed that clustering the unweighted mean word vector representation of RPs, i.e., a mean vector of the vectorized phrases' words, yielded inefficient concept separation, e.g., phrases ``American people'' and ``Mexican people'' were clustered into one concept although they refer to different nations. On the contrary, two phrases could be coreferential but only in the narrow event-determined context, e.g., ``young illegals'' - ``DACA recipients.'' 

To improve the effectiveness of clustering, we apply modifications to the vector representation, i.e., (1) employ a weighting scheme of the named entity (NE) components of the RPs and (2) calculate more than one similarity matrix to introduce more than one level of similarity between RPs. 

\subsubsection{Word vector weighting}

\begin{figure}[h]
\centering
\includegraphics[width=0.9\textwidth]{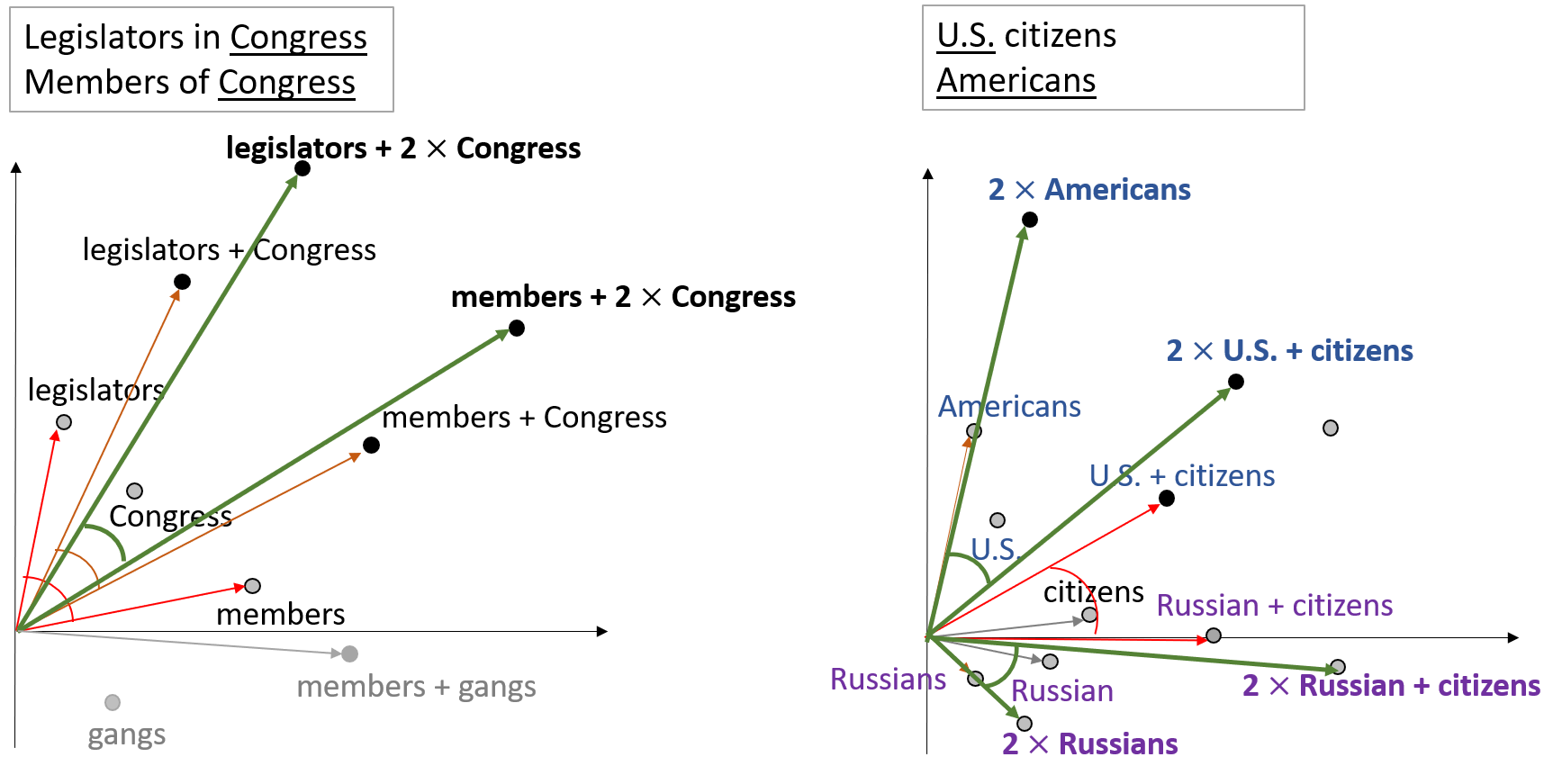}
\caption{\label{img:weight} The weighting of the NEs in phrases increases cosine similarity of related phrases and separates unrelated phrases.} 
\end{figure}

In the narrow article-specific context, word vector weighting \cite{zheng2015learning} increases the semantic proximity in the vector space and facilitates the identification of the semantic relatedness and coreferential relations (cf. Fig.~\ref{img:weight}). We represent phrases as the mean of their weighted words' embedding, i.e., 
\begin{flalign}
\label{weight}
V(rp_i) = \sum_{\forall i \in |rp|}^{}w_i \cdot v(i)
\end{flalign}
where $v(i)$ is a vector representation of the $i$-th word and $w_i$ is a weight assigned to this word. We use word2vec \cite{mikolov2013distributed} as a word embedding model due to its ability to represent both single words and multi-word phrases, resulting in more precisely defined positions of phrases in the vector space. 

A vector representation $V(rp_k)$ depends on its relations to $rp_l$ to which a similarity value is calculated. A weight $w_i$ for a word $v_i$ in (\ref{weight}) is selected as following:

\begin{flalign}
w_i = \begin{cases}
    NG_{ne(rp_{k}),ne(rp_{l})}, & \text{if } NG_{ne(rp_{k}), ne(rp_{l})} > 0 \\
    wt, &  \text{if } ne(rp_{k}) \in NG \text{ and } ne(rp_{l}) = \O \text{ or vice versa} \\ 
    1 ,& \text{else }
\end{cases}
\end{flalign}
where $ne(rp_i)$ is an extracted NE from $rp_i$, e.g., $ne($``Congress members''$)=$ ``\textit{Congress}'' (if $ne_t \notin rp_k 	\Rightarrow ne(rp_k) = \O$), $NG$ is a \textit{named entity (NE) grid}, i.e., a controlling matrix that allows or restricts similarity calculations between phrases that contain NEs, and $wt= 1.7$. 

An NE-grid $NG$ determines which types of mentions can be merged. For example, if $NG_{ne(rp_k), ne(rp_k)} = 0$, then the mentions of one geo-political entity (GPEs) are not compared to mentions of another GPEs, e.g., ``French'' $\neq$ ``North Korea.'' If a value of a NG's cell $NG_{ne(rp_k), ne(rp_l)} > 0$ then $NG$  favors to merge the corresponding RPs,  e.g., ``U.S.'' $=$ ``Americans.'' 

The NE-grid is spanned across combined NE chains $Ch$ of two types: country + nationality ($Ch_{cn}$) and organization + persons ($Ch_{op}$). To construct NE-chains, we use the relations between the terms in the semantic network ConceptNet \cite{speer2017conceptnet}. We iterated over the extracted NEs and interlinked them if their corresponding ConceptNet terms have a ``SimilarTo'' relation. Afterward, we restore full connectivity between the sub-chains, i.e., the restored connectivity of the extracted ``United States''-``U.S.'' and ``U.S.''-``American'' chains yields a chain $ch_{a}$  ``United States''-``U.S.''-``American.'' 

Based on the NE-chains, we constructed the NE-grid $NG$:

\begin{flalign}
NG_{ne_{k},ne_{l}} = \begin{cases}
    wt, &\quad  \text{if } ne_{k} \in ch_{a} \land ne_{l}  \in ch_{a} \text{ where } ch_a \in Ch_m \\
    1, & \quad \text{if } ne_{k} \in Ch_{m} \land ne_{l} \notin Ch_{m}   \\
    0, & \quad \text{if } ne_{k} \in Ch_{m} \land ne_{l} \in Ch_{m}
\end{cases}
\end{flalign}
where $m=cn \lor op$.

\subsubsection{Multiple similarity levels} 
To create additional levels of similarity, we calculate three similarity matrices: 1) a \textit{head-similarity matrix} $SH$,   2) a \textit{phrase-similarity matrix} $SP$, and 3) a \textit{core-phrase similarity matrix} $SPC$:

\begin{flalign}
SH_{h_i,h_j} = \begin{cases}
    \operatorname{cossim}(v(h_i),v(h_j))       & \hspace{0.1em} \text{if } h_i \neq h_j \\
    0.5 & \hspace{0.1em} \text{if } h_i = h_j
\end{cases}
\end{flalign}
\begin{flalign}
SP(C)_{rp_i,rp_j} = \begin{cases}
    0, & \hspace{0.1em} \text{if } \operatorname{cossim}(V(rp_i),V(rp_j)) < thr_{sim_{rp}}  \\
    & \quad \text{ or } rp_i = rp_j  \\ 
    & \quad \text{ or }  NG_{ne(rp_i),ne(rp_i)} = 0 \\
    \operatorname{cossim}(V(rp_i),V(rp_j)),  & \hspace{0.1em} \text{else} \\
    \end{cases}
\end{flalign}
where $h_k = h(rp_{i})$ is the head of a phrase, e.g., $h($``Congress members''$) = $ ``\textit{members},'' $cossim$ is cosine similarity, $v(\cdot)$/$V(\cdot)$ is a vector representation of words or phrases, $thr_{sim_{rp}} = 0.4$ is a threshold for the minimum RP similarity, and $SPC$ is a subset matrix of the $SP$ with the RPs that are core-mentions.

The output of the preprocessing step consists of three similarity matrices ($SH$, $SP$, $SPC$) that represent similarity of RPs as to three levels and an NE-grid $NG$ that determines restriction rules for operations between mentions.

\subsection{Identification of the cluster cores}
We start clustering with identification of the \textit{cluster cores} ($CC$), i.e., cluster the core mentions' RPs (CRP) as the most distinctive among all RPs (see Sec.~\ref{sec:extraction}). Two core RPs $crp_i$ and $crp_j$ form a $CC$ if they meet two requirements: (1) $SPC_{crp_i,crp_j} > 0$ and $SH_{crp_i,crp_j} > 0$, (2) $crp_i$ and $crp_j$ were similar to a sufficient number of other core RPs according to the \textit{ratio matrix} $RM$. 
Following OPTIC's principle of creating more similarity levels compared to one similarity metric, we form a \textit{ratio matrix} $RM$ for the core RPs. Each element in RM shows a normalized count of the core RPs to which two RPs at a hand are similar to: 
\begin{flalign}
RM_{crp_i,crp_j} = \begin{cases}
    frac & \text{if frac} \geq OR_{thr} \land crp_i \neq crp_j\\
    0 & \text{else}
\end{cases}
\end{flalign}
where
\begin{flalign}
frac = \frac{\sum (b(SPC_{crp_i, }) \land b(SPC_{crp_j, }))}{max(\sum b(SPC_{crp_i, }), \sum b(SPC_{crp_j}))} 
\end{flalign}
and $b(\cdot)$ is a binary representation of values in a vector (1 if a cell value is larger than 0, else 0);   $OR_{thr}=0.5 \leq \log_{5000} |RP| \leq 0.7$, i.e., the threshold is balanced based on the size of unique RPs: a larger number of RPs imposes more strict similarity requirements for the cluster cores.

\begin{figure}[h]
\centering
\includegraphics[width=8cm]{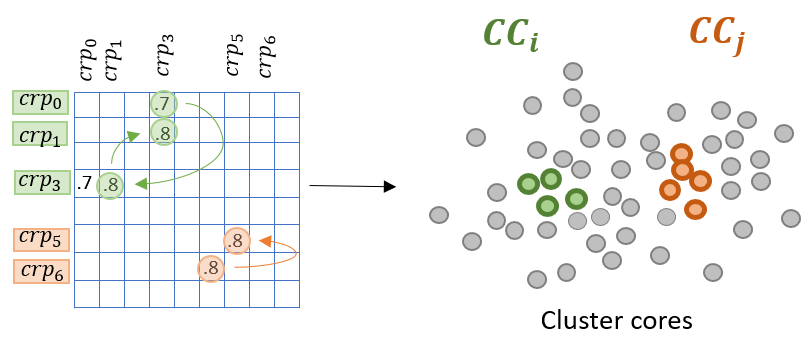}
\caption{\label{img:cores} Identification of chains of related core representatives: this example yields two core clusters.} 
\end{figure}

Finally, we iterate over the elements of $RM$ and recursively collect chains of the interlinked CRPs, as shown in Fig.~\ref{img:cores}. A chain is considered complete once no other core RPs can be added to it.

\subsection{Forming of cluster bodies}
\begin{figure}[h]
  \centering
  \begin{minipage}[t][][b]{0.45\textwidth}
    \includegraphics[height=4cm]{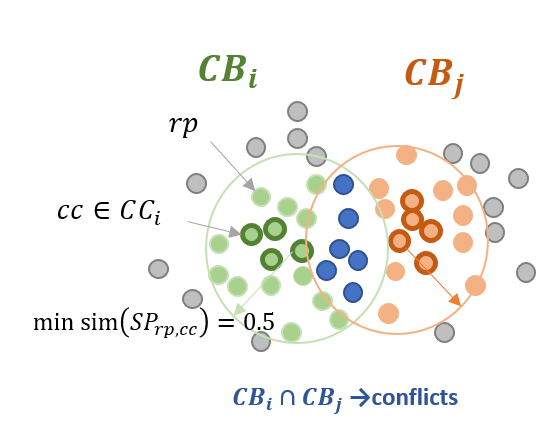}
    \caption{Identification of cluster bodies.}
    \label{img:bodies}
  \end{minipage}
  \hfill
  \begin{minipage}[t][][b]{0.45\textwidth}
    \includegraphics[height=4cm]{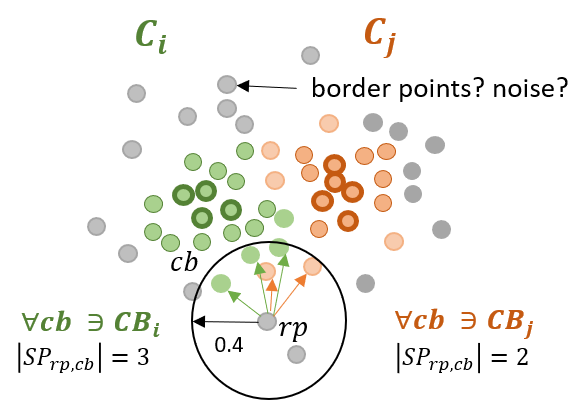}
    \caption{Adding border mentions.}
    \label{img:borders}
  \end{minipage}
\end{figure}

To further extend the clusters, we form \textit{cluster bodies} $CB$ by expanding the identified cored with the unclustered RPs (Fig.~\ref{img:bodies}). First, we assign RPs to the cluster cores if a RP was similar to at least one of the core RPs and the merge is allowed by $NG$:
\begin{multline}
CB_i = \{rp \cup CC_i| \; \forall rp \in RP, \exists cc \in CC_i: \\ 
SP_{rp, cc} \geq 0.5 \text{ and } NG_{ne(rp), ne(\forall CC_i)} \neq 0 \}
\end{multline}

Second, we intersect cluster bodies (CB) with each other to check if there were non-core RPs that belonged to both CBs. If so, we resolve the conflicting RPs by calculating a normalized similarity score between an $rp_{conf} \in CB_i \cap CB_j$ and non-conflicting RPs of each CB, and choosing a CB with the largest similarity score: 
\begin{flalign}
sim_{rp_{conf}, CB_i} = \frac{1}{|CB_i|} (\sum_{cb \in CB_i} |rp_{conf} \cap cb| +  \sum_{cb \in CB_i} SP_{rp_{conf}, cb})
\end{flalign}
\begin{flalign}
CB_{best} = \operatorname{arg\, max}_{i \in |CB|} sim_{rp_{conf}, CB_i}
\end{flalign}
i.e., similarity consists of the number of overlapping words between an RP and clustered RPs and the sum of their cross-similarity values.

\subsection{Adding border mentions} 
We define \textit{border mentions} as the remaining RPs that are similar at least to two body RPs (Fig. \ref{img:borders}). We add a border RP $rp$ to a cluster body $CB_i$ and formed a cluster $C_i$  if $rp$ is similar to at least two RPs in $CB_i$ and has the largest normalized similarity score to $CB_i$: 
\begin{multline}
C_i = \{rp \cup CB_i| \; \forall rp \in RP: \\
|SP_{rp, \forall cb \in CB_i} > 0| \geq 2 \quad \land \quad  NG_{ne(rp), ne(\forall cb \in CB_i)} \neq 0
\quad \land \\  max_{CB_i \in CB}(\frac{\sum_{cb \in CB_i} SP_{rp, cb}}{|\{\forall cb \in CB_i: SP_{rp, cb} > 0\}|})\}
\end{multline}

\subsection{Form non-core clusters} 
Some unmerged RPs can form non-core clusters, i.e., they are similar to other RPs but do not meet requirements to become core points (see Fig.\ref{img:pipeline_graph}). We form a \textit{non-core cluster} around a $rp$ as:
\begin{flalign}
nC_i = \{rp \bigcup_{rp_j \notin C} rp_j, \text{if } SP_{rp, rp_j} \geq 0.5\}
\end{flalign}

\subsection{Merging final clusters}
When all clusters are formed, the final step of the pipeline is to check if clusters can be further merged based on combined features of word count and word embeddings. We create an extended list of modifiers, i.e., all the previous (see Sec.\ref{sec:prep}) and also number and apposition modifiers. We compare the identified clusters according to a cosine similarity of the weighted vector representation using this extended list.

Each cluster $C_i$ is, first, represented with the counted RPs' lowercased lemmas  $L_i$. We treat clusters as documents and transformed the clusters into the TF-IDF representation \cite{zheng2015learning}. Each cluster $C_i$ is represented as a TF-IDF-weighted average word embedding representation of its lemmas:
\begin{flalign}
VC(C_i) = \frac{\sum_{l \in L_i} t(l) \cdot v(l)}{|L_i|}
\end{flalign}
where $t(l)$ if a TF-IDF coefficient of a lemma l in a cluster $C_i$.
We construct a cluster cross-similarity matrix $SC$, where each element  is:
\begin{multline}
SC_{C_i,C_j} = \begin{cases}
    sim & \text{if } sim \geq 0.6 \; \land  C_i \; \neq C_j \land \; \\
    & \quad   \forall l_k \in C_i \text{, } \forall l_l \in C_j:  NG_{ne(l_k), ne(l_l)} \neq 0 \\ 
    0  & \text{else}
\end{cases}
\end{multline}
where $sim = \operatorname{cossim}(VC(C_i),VC(C_j))$.

Following the principle from Fig.~\ref{img:cores}, we identify chains of clusters, i.e., the final clusters that contain related mentions.

\section{Preliminary evaluation and Discussion}

As a preliminary evaluation, we extracted concepts of (in)directly related mentions from five sets of event-related news articles with the identical parameters and we qualitatively analyzed the results. We used NewsWCL50 (N) \cite{Hamborg2019a} and ECB+ (e) \cite{cybulska2014using} as datasets that fulfill such criterion for the text collection. 

Table~\ref{tab:res} depicts examples of the identified concepts, i.e., clusters of the related mentions, from a subset of the events of each dataset. The column with concept names contains manually created labels that summarized automatically identified clusters of the related mentions. The column ``Mentions'' contains unique mentions of an identified clusters. Mentions are separated with the keywords that indicate the stages at which the mentions were clustered. 

The analysis of the indirectly referring mentions to groups of people shows that the proposed clustering approach successfully separated mentions related to GPEs such as ``Israeli officials'' and ``American officials.'' These mentions refer to different concepts but are quite similar due to the shared word ``officials.'' The identified concepts from the event N9 (``American officials,'' ``Iranian regime,'' ``Israeli officials,'', and ``European leaders'') show that the approach effectively separated mentions of multiple GPEs from the same text. 

Clustering of directly referring mentions, e.g., from the ``Central American migrants'' concept from event N6, resolves mentions such as ``Central American transgender women,'' ``asylum-seekers,'' ``caravan,'' and ``undocumented immigrants.'' This demonstrates that the proposed approach successfully clustered mentions that are exposed to context-specific coreference relations, i.e., none of these mentions are common-known synonyms to each other. Moreover, the approach successfully separated the ``Immigration lawyers'' concept from the ``Migrants'' concept although the noun ``immigration'' is shared among the two, which makes these mentions semantically similar. On the contrary, the ``Migrants'' concept contains falsely clustered mentions that refer to the various supporters of the immigrant caravan. Separation of such mentions with semantically close yet conceptually different meanings remains the biggest challenge for the algorithm and requires improvements to the clustering approach.

 \begin{table}[]
\begin{tabular}{p{0.037\textwidth}|p{0.2\textwidth}|p{0.75\textwidth}}
eID              & Concept name                                          & Mentions \\
\hline
\multirow{3}{*}{N1}  & Republican Congressional officials & CORE: House Republican committee chairmen, congressional committees, Republican chairmen, Republican Congressional intelligence officials, Congressional leadership BODY: House committees, congressional leaders, congressman BORDER: top aides, secretary, prudent law enforcement official, Leadership, chairmen, aides, his administration \\
                    \cline{2-3}
                    & Lawmakers                                & CORE: Select lawmakers, lawmakers, Many Democrats, analysts BODY: Conservatives   \\
                    \cline{2-3}
                    & Mueller investigators                           & CORE: investigators, Mueller investigators BODY: Federal prosecutors  \\
\hline
\multirow{2}{*}{N3}  & Russian agents                     & CORE: Russian agents, Russian intelligence agents, Russians BORDER: voters, its agents, Russian officials  \\
                    \cline{2-3}
                    & U.S. intelligence                     & CORE: American public, intelligence committees, American people, Americans, U.S. intelligence community BORDER: people, public  \\
\hline
\multirow{2}{*}{N6}  & Migrants                       & CORE: Central American migrants, asylum-seekers, Similar migrant groups, Central Americans, gay migrants, American sponsors, Central American children, several American advocacy groups, Asylum-seeking immigrant, Central American transgender women, refugees, their case, undocumented immigrants, immigrant rights activists BODY: Asylum-seekers, individuals, queer, migrant families, legitimate asylum-seekers, Migrant caravan, migrants, individual BORDER: caravan main organizing group, past 24-hours several groups, asylum seekers, families, his case, smugglers, immigration judges, particular group, caravan, sponsor, several groups, American sponsor, nonprofit group, children, Migrants, groups, protesters, his children, many migrants, group, their cases, her children, Immigrants, activists, their children, immigrants \\
                    \cline{2-3}
                    & Immigration lawyers                             & CORE: volunteer lawyers, good attorneys, volunteer attorneys, immigration lawyers BODY: legal observers BORDER: attorney  \\
                    \cline{2-3}
                    & U.S. authorities & CORE: U.S. government officials, Trump administration, U.S. authorities, U.S. immigration officials, American border authorities BODY: Southwest border states, Other administration officials BORDER: officer, authorities, officials, U.S. immigration lawyers, asylum officer, inspectors, administration, lawyers, U.S. families, Attorneys, credible-fear officers, Lawyers, his family, your family, international residents, his administration \\
\hline
\multirow{4}{*}{N9}  & American officials                   & CORE: Former intelligence officials, American officials, White House officials, outside experts, Officials BODY: Trump administration, intelligence community, officials BORDER: administration \\
                    \cline{2-3}
                    & Iranian regime                              & CORE: brutal regime, Iran leaders, exhaustive regimes, inspectors, inspection regime, Iranian regime BORDER: regime   \\
                    \cline{2-3}
                    & Israeli officials                               & CORE: senior Israeli official, Israelis, Israeli networks, Israeli leader, Israeli officials \\
                    \cline{2-3}
                    & European leaders                                & CORE: Europeans, European leaders  \\
\hline
\multirow{3}{*}{e41} & South Sudanese refugee camp                     & CORE: Yida camp, camp, Enough Project sources, South Sudanese refugee camp, sources, Yida refugee camp BORDER: refugee camp \\
                    \cline{2-3}
                    & South Sudan Liberation Army rebel group         & CORE: armed dissident groups, South Sudan Liberation Army rebel group, pro-southern groups, activist group, backing rebel groups, armed groups, minority ethnic group, American activist BORDER: their groups, group \\
                    \cline{2-3}
                    & Reuters correspondent                           & CORE: press conference, reporters, Reuters correspondent, November press conference BORDER: our correspondent
\end{tabular}
\caption{\label{tab:res}Results produced with the proposed concept identification approach. ``N''/``e''+ID indicates a dataset and the internal ID of the events of each dataset.}
\end{table}

\begin{table}[]
\begin{tabular}{p{0.037\textwidth}|p{0.2\textwidth}|p{0.75\textwidth}}
eID               & Concept name   & Mentions \\
\hline
\multirow{12}{*}{N6} & cl\_7  & Central American migrants, Central American children, several American advocacy groups, past 24-hours several groups, Other administration officials\\
                    \cline{2-3}
                    & migrants  & asylum-seekers, gay migrants, refugees, undocumented immigrants, Asylum-seekers, migrants, asylum seekers, smugglers, Migrants, Immigrants, immigrants\\
                    \cline{2-3}
                    & groups  & Similar migrant groups, caravan main organizing group, several groups, groups, protesters, group, activists \\
                    \cline{2-3}
                    & American sponsors & American sponsors, sponsor, American sponsor \\
                    \cline{2-3}
                    & Immigration lawyers  & Asylum-seeking immigrant, U.S. immigration lawyers, volunteer lawyers, volunteer attorneys, immigration lawyers  \\
                    \cline{2-3}
                    & case & their case, his case, their cases\\
                    \cline{2-3}
                    & cl\_20 & migrant families, families, children, his children, her children, their children, U.S. families, his family \\
                    \cline{2-3}
                    & cl\_0  & nonprofit group, many migrants, international residents \\
                    \cline{2-3}
                    & U.S. authorities & U.S. government officials, U.S. authorities, American border authorities, authorities, officials, inspectors, legal observers \\
                    \cline{2-3}
                    & Asylum officers & officer, asylum officer, credible-fear officers \\
                    \cline{2-3}
                    & Lawyers & lawyers, Attorneys, Lawyers, good attorneys, attorney \\
                    \cline{2-3}
                    & NOT    & Central Americans, Central American transgender women, immigrant rights activists, immigration judges, individuals, individual, queer, legitimate asylum-seekers, Migrant caravan, caravan, particular group, your family, Trump administration, U.S. immigration officials, Southwest border states, administration, his administration \\
\hline
\multirow{5}{*}{N9}  & officials  & American officials, White House officials, outside experts, Officials, officials, Israeli officials\\
                    \cline{2-3}
                    & regime  & administration, brutal regime, exhaustive regimes, Iranian regime, regime   \\
                    \cline{2-3}
                    & leaders  & Iran leaders, Israeli leader, European leaders \\
                    \cline{2-3}
                    & cl\_4  & senior Israeli official, Israelis, Europeans  \\
                    \cline{2-3}
                    & NOT    & Former intelligence officials, Trump administration, intelligence community, Israeli networks, inspectors, inspection regime                                                                                                                                                                                                            
\end{tabular}
\caption{Concepts identified by hierarchical clustering from the similar mentions of N6 and N9 in Table~\ref{tab:res}. The concepts are more narrowly defined or contain conceptually unrelated mentions. A lot of mentions compared to the proposed approach remain unclustered (``NOT'' cluster).  }
\label{tab:hc}
\end{table}

To test, if a state-of-the art clustering algorithm achieved similar concepts, we reclustered the mentions from two exemplary chosen documents, N6 and N9 in Table~\ref{tab:res}, with hierarchical clustering (HC). Table~\ref{tab:hc} shows the results of HC with average linkage criterion, cosine distance (using a threshold 0.7) for both datasets.\footnote{The threshold was optimized per event as the one producing both the highest mean cross-phrase cosine similarity and clustering the most phrases.} Likewise in Table~\ref{tab:res}, we manually named the concepts which contained conceptually related mentions.  While some of the mentions formed more narrowly and fine-grained defined concepts, HC also clustered conceptually different mentions and left approximately 25\% of the input mentions unclustered (``NOT'' clusters in Table~\ref{tab:hc}).

The proposed clustering approach might be beneficial to cross-document coreference resolution (CDCR), i.e., resolution of the coreferential mentions of various entities across sets of related text documents.  Such entity types as groups of people and mentions of the GPEs are some of the targets for CDCR. When implemented as a part of a CDCR model, our concept identification approach can have strong positive impact to the overall performance due the resolution of coreferential mentions of high lexical diversity. Such mentions are typically a subject of bias of word choice and labeling, i.e., contain biased wording that contains polarized connotation and typically is coreferential only in a narrow context of a reported event.

\section{Conclusion and Future work}
We proposed a clustering approach to identify both direct mentions referring to groups of individuals and indirect person mentions related to the geo-political entity (GPEs) or organizations, i.e., job titles that represent these entities. In our evaluation, we found that terms such as ``American officials'' were resolved reliably as mentions related to GPEs or organizations. Moreover, the approach capably clustered mentions that lack NE-components while maintaining a fine-grained level of conceptualization among the clusters of these mentions. Further, the approach resolved mentions referring to groups of individuals that have highly-context dependent synonymous or coreferential relations, as apposed to universal synonyms. Thus, we think the approach is a robust solution to cross-document coreference resolution (CDCR), especially when employed in texts containing coreferential mentions with high lexical diversity.

As future work directions, we seek to test the proposed approach with other word vector models, e.g., fastText \cite{mikolov2018advances} and ELMo \cite{peters2018deep}, or phrase vector models \cite{wu2020phrase2vec}, pre-trained and fine-tuned on event-related news articles. We also seek to address current shortcomings, e.g., to resolve one-word mentions without modifiers, e.g., ``officials,'' we plan to devise an additional word sense disambiguation step. Each particular occurrence of a one-word mention will be resolved based on the mention's the context. Lastly, we will perform a quantitative analysis of the approach applied to CDCR, i.e., tested on the state-of-the art manually annotated CDCR datasets.

\bibliographystyle{splncs04}
\bibliography{ms}

\begin{thebibliography}{10}
\providecommand{\url}[1]{\texttt{#1}}
\providecommand{\urlprefix}{URL }
\providecommand{\doi}[1]{https://doi.org/#1}

\bibitem{optics}
Ankerst, M., Breunig, M.M., Kriegel, H.P., Sander, J.: Optics: Ordering points
  to identify the clustering structure. In: Proceedings of the 1999 ACM SIGMOD
  International Conference on Management of Data. p. 49–60. SIGMOD ’99,
  Association for Computing Machinery, New York, NY, USA (1999).
  \doi{10.1145/304182.304187}, \url{https://doi.org/10.1145/304182.304187}

\bibitem{cambria2018senticnet}
Cambria, E., Poria, S., Hazarika, D., Kwok, K.: Senticnet 5: Discovering
  conceptual primitives for sentiment analysis by means of context embeddings.
  In: Thirty-Second AAAI Conference on Artificial Intelligence (2018)

\bibitem{cha2017language}
Cha, M., Gwon, Y., Kung, H.: Language modeling by clustering with word
  embeddings for text readability assessment. In: Proceedings of the 2017 ACM
  on Conference on Information and Knowledge Management. pp. 2003--2006 (2017)

\bibitem{chen2018anchorviz}
Chen, N.C., Suh, J., Verwey, J., Ramos, G., Drucker, S., Simard, P.: Anchorviz:
  Facilitating classifier error discovery through interactive semantic data
  exploration. In: 23rd International Conference on Intelligent User
  Interfaces. pp. 269--280 (2018)

\bibitem{cybulska2014using}
Cybulska, A., Vossen, P.: Using a sledgehammer to crack a nut? lexical
  diversity and event coreference resolution. In: LREC. pp. 4545--4552 (2014)

\bibitem{Hamborg2019a}
Hamborg, F., Zhukova, A., Gipp, B.: Automated identification of media bias by
  word choice and labeling in news articles. In: Proceedings of the ACM/IEEE
  Joint Conference on Digital Libraries (JCDL) (Jun 2019).
  \doi{10.1109/JCDL.2019.00036}

\bibitem{Hamborg2019}
Hamborg, F., Zhukova, A., Gipp, B.: Illegal aliens or undocumented immigrants?
  towards the automated identification of bias by word choice and labeling. In:
  Proceedings of the iConference 2019 (Mar 2019).
  \doi{10.1007/978-3-030-15742-5\_17}

\bibitem{han2017automatic}
Han, X., Wu, Z., Huang, P.X., Zhang, X., Zhu, M., Li, Y., Zhao, Y., Davis,
  L.S.: Automatic spatially-aware fashion concept discovery. In: Proceedings of
  the IEEE International Conference on Computer Vision. pp. 1463--1471 (2017)

\bibitem{JIA2018691}
Jia, C., Carson, M.B., Wang, X., Yu, J.: Concept decompositions for short text
  clustering by identifying word communities. Pattern Recognition  \textbf{76},
   691 -- 703 (2018). \doi{https://doi.org/10.1016/j.patcog.2017.09.045},
  \url{http://www.sciencedirect.com/science/article/pii/S0031320317303953}

\bibitem{kim2017bag}
Kim, H.K., Kim, H., Cho, S.: Bag-of-concepts: Comprehending document
  representation through clustering words in distributed representation.
  Neurocomputing  \textbf{266},  336--352 (2017)

\bibitem{manning-EtAl:2014:P14-5}
Manning, C.D., Surdeanu, M., Bauer, J., Finkel, J., Bethard, S.J., McClosky,
  D.: The {Stanford} {CoreNLP} natural language processing toolkit. In:
  Association for Computational Linguistics (ACL) System Demonstrations. pp.
  55--60 (2014), \url{http://www.aclweb.org/anthology/P/P14/P14-5010}

\bibitem{mikolov2018advances}
Mikolov, T., Grave, E., Bojanowski, P., Puhrsch, C., Joulin, A.: Advances in
  pre-training distributed word representations. In: Proceedings of the
  International Conference on Language Resources and Evaluation (LREC 2018)
  (2018)

\bibitem{mikolov2013distributed}
Mikolov, T., Sutskever, I., Chen, K., Corrado, G.S., Dean, J.: Distributed
  representations of words and phrases and their compositionality. In: Advances
  in neural information processing systems. pp. 3111--3119 (2013)

\bibitem{murtagh2012algorithms}
Murtagh, F., Contreras, P.: Algorithms for hierarchical clustering: an
  overview. Wiley Interdisciplinary Reviews: Data Mining and Knowledge
  Discovery  \textbf{2}(1),  86--97 (2012)

\bibitem{nikfarjam2015pharmacovigilance}
Nikfarjam, A., Sarker, A., O’connor, K., Ginn, R., Gonzalez, G.:
  Pharmacovigilance from social media: mining adverse drug reaction mentions
  using sequence labeling with word embedding cluster features. Journal of the
  American Medical Informatics Association  \textbf{22}(3),  671--681 (2015)

\bibitem{peters2018deep}
Peters, M.E., Neumann, M., Iyyer, M., Gardner, M., Clark, C., Lee, K.,
  Zettlemoyer, L.: Deep contextualized word representations. In: Proceedings of
  NAACL-HLT. pp. 2227--2237 (2018)

\bibitem{si2019enhancing}
Si, Y., Wang, J., Xu, H., Roberts, K.: Enhancing clinical concept extraction
  with contextual embeddings. Journal of the American Medical Informatics
  Association  \textbf{26}(11),  1297--1304 (2019)

\bibitem{speer2017conceptnet}
Speer, R., Chin, J., Havasi, C.: Conceptnet 5.5: An open multilingual graph of
  general knowledge. In: Thirty-First AAAI Conference on Artificial
  Intelligence (2017)

\bibitem{subramanian-roth-2019-improving}
Subramanian, S., Roth, D.: Improving generalization in coreference resolution
  via adversarial training. In: Proceedings of the Eighth Joint Conference on
  Lexical and Computational Semantics (*{SEM} 2019). pp. 192--197. Association
  for Computational Linguistics, Minneapolis, Minnesota (Jun 2019).
  \doi{10.18653/v1/S19-1021}, \url{https://www.aclweb.org/anthology/S19-1021}

\bibitem{wu2020phrase2vec}
Wu, Y., Zhao, S., Li, W.: Phrase2vec: Phrase embedding based on parsing.
  Information Sciences  \textbf{517},  100--127 (2020)

\bibitem{zheng2015learning}
Zheng, G., Callan, J.: Learning to reweight terms with distributed
  representations. In: Proceedings of the 38th international ACM SIGIR
  conference on research and development in information retrieval. pp. 575--584
  (2015)

\end{thebibliography}

\end{document}